\documentclass[a4paper]{article}

\usepackage{INTERSPEECH2018}
\usepackage{multirow}

\newcommand*{\sigmoid}{\operatorname{sigmoid}}

\title{Towards Unsupervised Automatic Speech Recognition \\ Trained by Unaligned Speech and Text only}
\name{Yi-Chen Chen, Chia-Hao Shen, Sung-Feng Huang, Hung-yi Lee}
\address{National Taiwan University, Taiwan}
\email{ \{r06942069, r04921047, r06942045, hungyilee\}@ntu.edu.tw }

\begin{document}

\maketitle
% 
%%%%%%%%% ABSTRACT
\begin{abstract}
%我建議不要用 weakly-supervised ，過去只用少量 alignment 的 paper 也都說是 weakly-supervised，這樣很容易造成誤解，就直接說是 unsupervised -- Lee

Automatic speech recognition (ASR) has been widely researched with supervised approaches, while many low-resourced languages lack audio-text aligned data, and supervised methods cannot be applied on them. 
  In this work, we propose a framework to achieve unsupervised ASR on a read English speech dataset, where audio and text are unaligned. 
In the first stage, each word-level audio segment in the utterances is represented by a vector representation extracted by a sequence-of-sequence autoencoder, in which phonetic information and speaker information are disentangled.
 Secondly, semantic embeddings of audio segments are trained from the vector representations using a skip-gram model. 
Last but not the least, an unsupervised method is utilized to transform semantic embeddings of audio segments to text embedding space, and finally the transformed embeddings are mapped to words.
  With the above framework, we are towards unsupervised ASR trained by   unaligned text and speech only.

\end{abstract}

%%%%%%%%% INTRODUCTION
\section{Introduction}

%ASR
ASR has reached a huge success and been widely used in modern society~\cite{cooke2001robust,povey2011kaldi,sha2007large}.
However, in the existing algorithms, machines must learn from a large amount of annotated data, which makes the development of speech technology for a new language with low resource challenging. 
%Therefore, for low-resource languages with scarce annotated data or languages without written forms, sufficiently accurate speech recognition is difficult to achieve. 
Annotating audio data for speech recognition is expensive, but unannotated audio data is relatively easy to collect.
If the machine can acquire the word patterns behind speech signals from a large collection of unannotated speech data without alignment with text, it would be able to learn a new language from speech in a novel linguistic environment with little supervision. 
%Imagine a Hokkien-speaking family buying an intelligent device: although at first the machine does not understand Hokkien, by hearing people speak it, it automatically learns the language. 
There are lots of researches towards this goal~\cite{Zero2017,Gish06icassp,pattern_IS11,SDTW-pattern,pattern_NMF,pattern_NMF_beta,Wang12icassp,SegmentASRU2017}. 
%藉此引用大量前人論文 %如果因此導致 reference 太長，以致空間不夠，優先拿掉我們實驗室的文章，還是不行，如果有些作者的文章有好幾篇，就保留最新的-- Lee

%Audio Word2Vec
Audio  segment representation is still an open problem with lots of research~\cite{SRAILICASSP15,WordEmbedIS14,QbyELSTMICASSP15,settle2017query,SemanticRepresentationICASSP18}.
In the previous work, a sequence-to-sequence autoencoder (SA) is used to represent variable-length audio segments using fixed-length vectors~\cite{chung2016audio,shen2017audio}.
In SA, the RNN encoder reads an audio segment represented as an acoustic feature sequence and maps it to a vector representation; the RNN decoder maps the vector back to the input sequence of the encoder. %The RNN encoder and decoder are trained to minimize the reconstruction error of the input acoustic sequence. 
With SA, only audio segments without human annotation are needed, which suits it for low-resource applications. 
%Moreover, the sequence-to-sequence learning~\cite{sutskever2014sequence} can handle input and output with arbitrary length, and has been applied in natural language processing and video processing.
It has been shown that the vector representation contains phonetic information~\cite{chung2016audio,shen2017audio,GANfeatureASRU17,hsu2017unsupervised}.

%看不懂
%In addition, because speaker and environmental information in original audio can degrade the performance of ASR, the techniques for learning of disentangled and interpretable representations from SA~\cite{GANfeatureASRU17,hsu2017unsupervised} can be very useful for extracting speaker invariant representations.
%However, although the audio segments of the same words may have close representations from SA, the recognition of words with similar pronunciation often fails in traditional ASR.
%Humans also encounter the same problem while hearing similar words.
%To deal with it, humans often consider contextual information to distinguish these words.
%Inspired from this, if we can obtain semantic embeddings, which contain rich contextual information, from representations extracted from SA, we believe the above problem of ASR can be alleviated.

%semantic
In text, Word2Vec~\cite{mikolov2013distributed} transforms each word into a fixed-dimension semantic vector used as the basic component of applications of natural language processing.
Word2Vec is useful because it is learned from a large collection of documents without supervision.
In this paper, we propose a similar method to extract semantic representations from audio without supervision. %and further transform it to text. 
First, phonetic embeddings from audio segments with little speaker or environment dependent information are extracted by SA with adversarial training for disentangling information. 
Then, the phonetic embeddings are further used to obtain semantic embeddings by a skip-gram model~\cite{mikolov2013distributed}.
%Rather than taking one-hot representations of words as input as in text, here we use phonetic embeddings from SA as input of the model.
Different from typical Word2Vec which takes  one-hot representations of words as input, here the proposed model takes phonetic embeddings from SA as input.

%Given a set of word embeddings learned from text, the text corresponding to the semantic embeddings of the audio segments would be available if we can map the audio semantic embeddings to the textual semantic embedding space.
Given a set of word embeddings learned from text, if we can map the audio semantic embeddings to the textual semantic embedding space, the text corresponding to the semantic embeddings of the audio segments would be available.
In this way, unsupervised ASR would be achieved.
The idea is inspired from unsupervised machine translation with monolingual corpora only~\cite{lample2017unsupervised,conneau2017word}.
Because most languages share the same expressive power and are used to describe similar human experiences across cultures, they should share similar statistical properties. For example, one can expect the most frequent words to be shared.
Therefore, given two sets of word embeddings of two languages, the representations of these words can be similar up to a linear transformation~\cite{conneau2017word,hoshen2018iterative}.

In our task, the targets we want to align are not two different languages, but audio and text of the same language.
We believe the alignment is probable because the frequencies and contextual relations of words are close in audio and text domains for the same language.  
The mapping method used in this stage is an EM-based method, Mini-Batch Cycle Iterative Closest Point (MBC-ICP)~\cite{hoshen2018iterative}, which  is originally proposed for unsupervised machine translation. 
Here given two sets of embeddings, that is, semantic embeddings from text and audio, MBC-ICP can iteratively align the vectors in the two sets by Principal Component Analysis (PCA) and an affine transformation matrix.
%This statistical advantage make MBC-ICP a suitable alignment method between the audio and text domains.
After mapping the semantic embeddings from audio to those learned from text, the text corresponding to the audio segments is directly known. %and  unsupervised ASR is thus achieved.

To our best knowledge, this is the first work attempting to achieve word-level ASR without any speech and text alignment. %\footnote{There is a companion paper submitted to INTERSPEECH~\cite{}, which achieved unsupervised phoneme recognition.}.

%\footnote{The code will be soon released.}

%%%%%%%%% PROPOSED METHOD
\section{Proposed Method}

The proposed framework of unsupervised ASR consists of three stages: 
\begin{enumerate}
\item Extracting phonetic embeddings from word-level audio segments using SA with discrimination.
\item Training semantic embeddings from phonetic embeddings.
\item Unsupervised transformation from audio semantic embeddings to textual semantic embeddings.
\end{enumerate}
The above three stages will be described in Sections~\ref{subsec:stage1},~\ref{subsec:stage2} and~\ref{subsec:stage3} respectively.

%the first stage
\subsection{Extracting Phonetic Embeddings from Word-Level Audio Segments Using SA with Discrimination} \label{subsec:stage1}
% autoencoder.png 和 discriminator.png 應該合併為一張，我直接改在圖上 -- Lee
\begin{figure}[t]
  \centering
  \includegraphics[width=\linewidth]{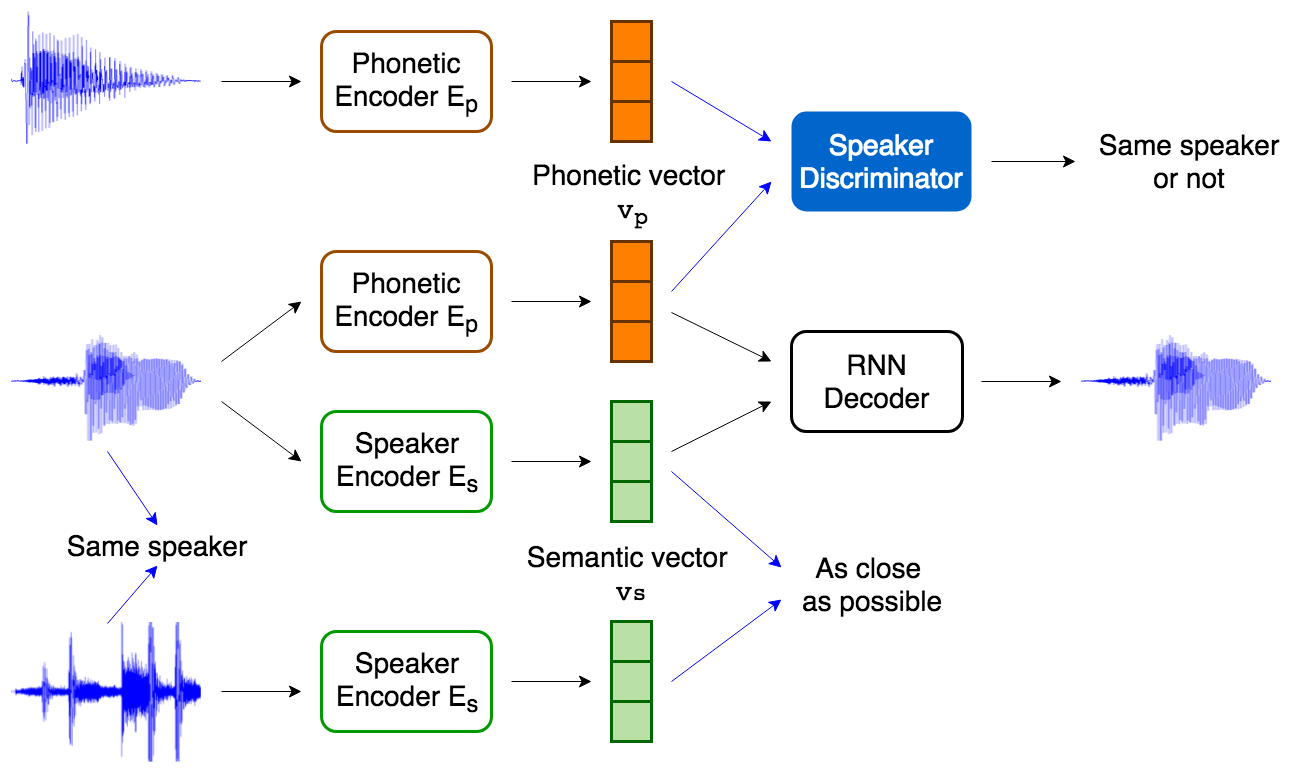}
  \caption{Network architecture for disentangling phonetic and speaker information.}
  \label{fig:autoencoder}
\end{figure}
%\begin{figure}[t]
%  \centering
%  \includegraphics[width=\linewidth]{discriminator.png}
%  \caption{Discriminator}
%  \label{fig:discriminator}
%\end{figure}

In the proposed framework, we assume that in an audio collection, each utterance is already segmented into word-level segments.
Although unsupervised segmentation is still challenging, there are already many approaches available~\cite{wang2017gate,scharenborg2010unsupervised}.
We denote the audio collection as $\mathbf{X} = {\{\mathbf{x}_{i}\}}_{i=1}^{M}$, which consists of $M$ word-level audio segments, $\mathbf{x}=(x_1, x_2, ..., x_T)$, where $x_t$ is the feature vector of the t\textsuperscript{th} time frame and $T$ is the number of time frames of the segment. 
%Each feature sequence corresponds to a word label $w \in W=\{w_1, w_2, ..., w_N\}$ and a speaker label $s \in S=\{s_1, s_2, ..., s_M\}$, where $N$ is the number of all words and $M$ is the number of speakers. 
The goal is to disentangle the phonetic and speaker information in acoustic features, and extract a vector representation with phonetic information.

\subsubsection{Autoencoder}
%The architecture consists of two parts: an autoencoder and a discriminator. 
As shown in Figure~\ref{fig:autoencoder}, we pass a sequence of acoustic features $\mathbf{x}$ into a phonetic encoder $E_p$ and a speaker encoder $E_s$ to obtain a phonetic vector $\mathbf{v_p}$ and a speaker vector $\mathbf{v_s}$. 
Then we take the phonetic and speaker vectors as inputs of the decoder to reconstruct the acoustic features $\mathbf{x}'$. 
The phonetic vector $\mathbf{v_p}$ will be used in the next stage.
The two encoders and the decoder are jointly learned by minimizing the reconstruction loss below:
\begin{equation}
\begin{aligned}
L_r &= \sum_{i} || \mathbf{x}_i - \mathbf{x}_i' ||_2. 
  \label{reconstruction_loss}
\end{aligned}
\end{equation} 
It will be clear in Sections~\ref{subsubsec:speaker} and~\ref{subsubsec:phonetic} how to make $E_p$ encode phonetic information and  $E_s$ encode speaker information.

\subsubsection{Training Criteria for Speaker Encoder} \label{subsubsec:speaker}
In the following discussion, we also assume the speakers of the segments are known.
Suppose the segment $\mathbf{x}_i$ is uttered by speaker $s_i$.
If speaker information is not available, we can simply assume that the segments from the same utterance are uttered by the same speakers, and the approach below can still be applied.
$E_s$ is learned to minimize the following loss $L_s$:
\begin{equation}
\begin{aligned}
L_s &= \sum_{s_i = s_j} || \mathbf{v_s}_i - \mathbf{v_s}_j ||_2 \\
  &+ \sum_{s_i \neq s_j}  \max( \lambda - ||  \mathbf{v_s}_i - \mathbf{v_s}_j ||_2, 0).
  \label{speaker_loss}
\end{aligned}
\end{equation} 
If $\mathbf{x}_i$ and $\mathbf{x}_j$ are uttered by the same speaker ($s_i = s_j$), we want their speaker embeddings $\mathbf{v_s}_i$ and $\mathbf{v_s}_j$ to be as close as possible.
On the other hand, if $s_i \neq s_j$, we want the distance of $\mathbf{v_s}_i$ and $\mathbf{v_s}_j$ larger than a threshold $\lambda$.
%We also add a speaker loss term to force the speaker vectors from the same speaker to be close, while speaker vectors coming from different speakers stay away from one another by at least a distance.

\subsubsection{Training Criteria for Phonetic Encoder} \label{subsubsec:phonetic}

As shown in Figure~\ref{fig:autoencoder}, the discriminator $D$ takes two phonetic vectors $\mathbf{v_p}_i$ and $\mathbf{v_p}_j$ as inputs and tries to tell if the two vectors come from the same speaker. 
The learning target of the phonetic encoder $E_p$ is to "fool" the discriminator, keeping it from discriminating correctly. 
In this way, only phonetic information is contained in the phonetic vector, and the speaker information in original acoustic features is encoded in the speaker vector. 
The discriminator learns to maximize $L_d$ in (\ref{discriminative_loss}), while the phonetic encoder learns to minimize $L_d$. 
\begin{equation}
\begin{aligned}
L_d &= \sum_{s_i = s_j} D(\mathbf{v_p}_i, \mathbf{v_p}_j) - \sum_{s_i \neq s_j} D(\mathbf{v_p}_i, \mathbf{v_p}_j).
  \label{discriminative_loss}
\end{aligned}
\end{equation} 
The whole optimization procedure of the discriminator and the other parts is iteratively minimizing $L_d$ and $L_r + L_s - L_d$.

%the second stage
\subsection{Training Semantic Embeddings from Phonetic Embeddings} \label{subsec:stage2}

\begin{figure}[t]
  \centering
  \includegraphics[width=\linewidth]{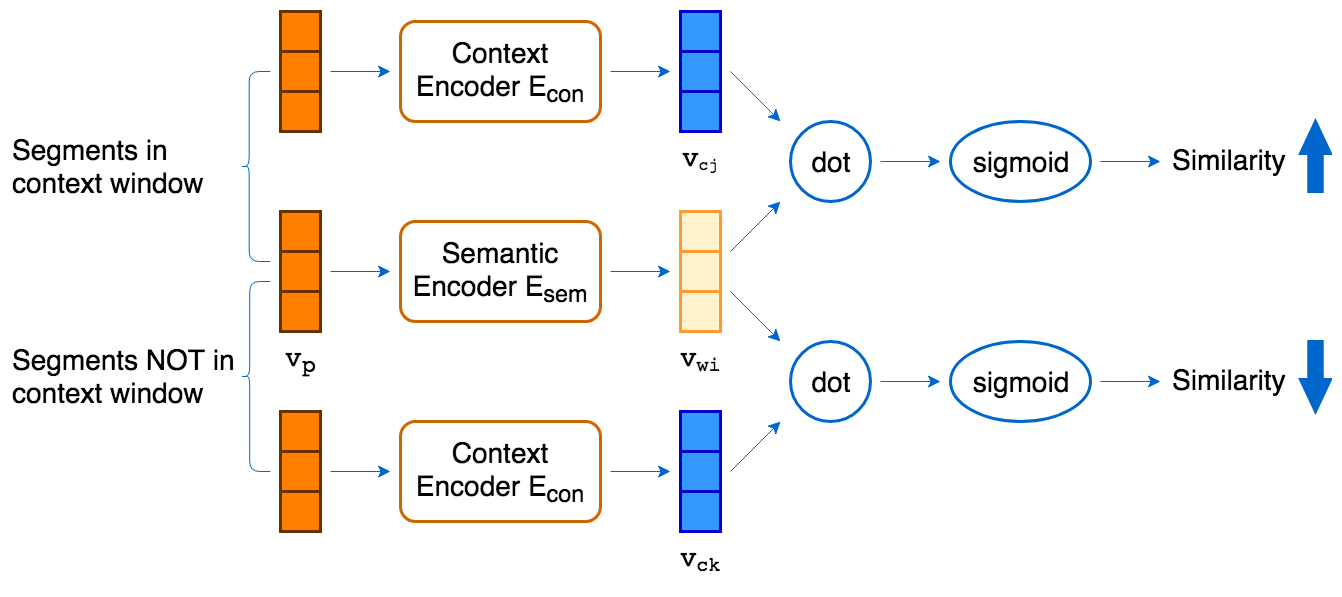}
  \caption{Training semantic embeddings from phonetic embeddings.}
  \label{fig:semantic}
\end{figure}

Similar to the Word2Vec skip-gram model~\cite{mikolov2013distributed}, we use two encoders $E_{\text{sem}}$ and $E_{\text{con}}$ to train the semantic embeddings from phonetic embeddings (Figure~\ref{fig:semantic}). %記得把 $E_{sem}$ and $E_{con}$ 標到圖上 -- Lee
On one hand, given a segment $\mathbf{x_i}$, we feed its phonetic vector $\mathbf{v_p}_i$ obtained from the previous stage into $E_{\text{sem}}$, and output the semantic embedding of the segment $\mathbf{v_w}_i = E_{\text{sem}}(\mathbf{v_p}_i) $. 
On the other hand, given the context window size $c$, which is a hyperparameter, if a segment $\mathbf{x_j}$ is in the context window of $\mathbf{x_i}$, then its phonetic vector $\mathbf{v_p}_j$ is a context vector of $\mathbf{v_p}_i$. 
For each context vector $\mathbf{v_p}_j$ of $\mathbf{v_p}_i$, we feed it into $E_{\text{con}}$, and output its context embedding $\mathbf{v_c}_j = E_{\text{con}}(\mathbf{v_p}_j)$.

Given a pair of phonetic vectors $(\mathbf{v_p}_i, \mathbf{v_p}_j)$, the training criteria for $E_{\text{sem}}$ and $E_{\text{con}}$ is to maximize the similarity of $\mathbf{v_w}_i$ and $\mathbf{v_c}_j$ if $\mathbf{v_p}_i$ and $\mathbf{v_p}_j$ are contextual, while minimizing their similarity otherwise.
The basic idea is parallel to textual Word2Vec.
Two different words having the similar content have similar semantics, thus if two different phonetic embeddings corresponding to different words have the same context, they will be close to each other after projected by $E_{\text{sem}}$. 
$E_{\text{sem}}$ and $E_{\text{con}}$ learn to minimize the semantic loss $L_{\text{sem}}$ as follows: 
\begin{equation}
\begin{aligned}
L_{\text{sem}} &=  \sum_{(\mathbf{x_i}, \mathbf{x_j}) \text{~in context window}} - \log(\sigmoid(\mathbf{v_w}_i \cdot \mathbf{v_c}_j)) \\
  &+ \sum_{(\mathbf{x_i}, \mathbf{x_k}) \text{~not in context window}} - \log(\sigmoid(- \mathbf{v_w}_i \cdot \mathbf{v_c}_k)). 
 %要加負號才是 loss
 %看看這樣對不對? -- Lee
  \label{semantic_loss}
\end{aligned}
\end{equation} 
The sigmoid of dot product of $\mathbf{v_w}$ and $\mathbf{v_c}$ is used to evaluate the similarity.
If $\mathbf{x_i}$ and $\mathbf{x_j}$ are in the same context window, we want $\mathbf{v_w}_i$ and $\mathbf{v_c}_j$ to be as similar as possible. 
We also use the negative sampling technique, in which only some pairs $(\mathbf{x_i},\mathbf{x_k})$ are randomly sampled as negative examples instead of enumerating all possible negative pairs.
%That is, we randomly choose another word $w'$ in training data, obtain its semantic embedding $\mathbf{v_w}'$, and let the dot product of $\mathbf{v_w}'$ and $\mathbf{v_w}$ be less than a threshold. %"less than a threshold"是哪來的?在上式看不出來 -- Lee
%In this way, the semantic embeddings are not all the same and are therefore meaningful. 不用這句

%多餘的
%Since $\mathbf{v_s}$ and $\mathbf{v_c}_j$ are contextual, we would like to maximize the log likelihood: 
%\begin{equation}
%\begin{aligned}
%\sum_{j} log\ p(\mathbf{v_c}_j | \mathbf{v_w}) 
%\end{aligned}
%\end{equation}

%the third stage
\subsection{Unsupervised Transformation from Audio to Text} \label{subsec:stage3}
% 這一段很有可能會被 reviewer 質疑 (尤其如果實驗結果沒有很好的話)，我感覺原來 FB 那篇也不是特別有說服力，但是因為結果很好，所以大家就算了 ..;
% 這是我的感覺，目前還沒想到特別好的解法 -- Lee

We have a set of audio semantic embeddings $\mathbf{V_W} = \{\mathbf{v_w}_1, ...,\mathbf{v_w}_i, ..., \mathbf{v_w}_M\}$ obtained from the last stage, where $M$ is the number of audio segments in the audio collection. %Here $\mathbf{v_w}_i$ is the semantic embeddings of audio segment 
On the other hand, given a text collection, we can obtain textual semantic embeddings $\mathbf{U_W} = \{\mathbf{u_w}_1, ...,\mathbf{u_w}_j, ...,\mathbf{u_w}_N\}$ by typical word embedding models like skip-gram. %as well. 
Here $\mathbf{u_w}_j$ is the word embedding of the $j$-th word in the text collection, and there are $N$ words in the text database.
Although both $\mathbf{v_w}$ and $\mathbf{u_w}$ contain semantic information, they are not in the same space, that is, the same dimension in $\mathbf{v_w}$ and $\mathbf{u_w}$ would not correspond to the same semantic meaning.
Here we want to learn a transformation to transform an embedding $\mathbf{v_w}$ to $\mathbf{\tilde{u}_w}$ in the textual semantic space. %where $\mathbf{v_w}$ and $\mathbf{\tilde{u}_w}$ correspond to the same word. 

%$\mathbf{\tilde{u}_w} = \mathbf{T}(\mathbf{v_w})$. 後面根本沒有 T 只有 T_xy, T_yx

%As long as we have the transformation $\mathbf{T}$, we can map the semantic embeddings $\mathbf{v_w}$ of all the audio segments to the textual space $\mathbf{\tilde{u}_w}$.
%Using a similar skip-gram model, we can obtain textual semantic embeddings $\mathbf{U_W} = \{\mathbf{u_w}_1, ..., \mathbf{u_w}_M\}$ as well. 
%Given audio semantic embeddings $\mathbf{V_W} = \{\mathbf{v_w}_1, ..., \mathbf{v_w}_M\}$, we want to transform $\mathbf{V_W}$ to $\mathbf{U_W}$ by an affine transformation $\mathbf{T}$ and vise versa. 

%非常重要！！！！！！　x　前面已經用過了， 代表 audio ，請務必換一下 notation (不是換這邊，就是換前面) -- Lee
MBC-ICP is used here, whose procedure is described as below. 
Given two sets of embeddings, $\mathbf{V_W}$ and $\mathbf{U_W}$, they are projected to their top $K$ principal components by PCA respectively. 
Let the projected vectors of $\mathbf{V_W}$ and $\mathbf{U_W}$ be $\mathbf{A}$ and $\mathbf{B}$.
The $i$-th column of $\mathbf{A}$, $\mathbf{a_i}$, is the PCA projection of $\mathbf{v_w}_i$, while the $j$-th column of $\mathbf{B}$, $\mathbf{b_j}$, is the PCA projection of $\mathbf{u_w}_j$.
Both the dimensionality of $\mathbf{a_i}$ and $\mathbf{b_j}$ are $K$.
If $\mathbf{V_W}$ can be mapped to the space of $\mathbf{U_W}$ by an affine transformation, $\mathbf{A}$ and $\mathbf{B}$ would be similar after PCA~\cite{hoshen2018iterative}.
The above PCA mapping technique is commonly used~\cite{daras2012investigating,li2017method}. 
%Assume that ``politics'' is the most frequent topic mentioned in both audio and text database.
%Although it may lies in different dimensions in $\mathbf{v_w}$ and $\mathbf{u_w}$, it would corresponds to similar principal components after PCA.
%Therefore, in $\mathbf{a}$ and $\mathbf{b}$ both the first dimension (for the strongest component) corresponds to ``politics''.
%這可能不是最好的說明，歡迎修改 -- Lee
%Although the distributions of $\mathbf{V_W}$ and $\mathbf{U_W}$ are different, after whitening by PCA, they can be aligned together.

Then a pair of transformation matrices, $\mathbf{T_{ab}}$ and $\mathbf{T_{ba}}$, is learned, where $\mathbf{T_{ab}}$ transforms an $\mathbf{a}$ in $\mathbf{A}$ to the space of $\mathbf{B}$, that is, $\tilde{\mathbf{b}} = \mathbf{T_{ab}}\mathbf{a}$, while $\mathbf{T_{ba}}$ maps $\mathbf{b}$ to the space of $\mathbf{A}$.
$\mathbf{T_{ab}}$ and $\mathbf{T_{ba}}$ are learned iteratively by the following algorithm.
We assume that two kinds of semantic embedding are likely the same after PCA projection, so we initialize the transformation matrices as identity matrices. 
Then in each iteration, the following steps are conducted:
\begin{enumerate}
\item For each $\mathbf{a_i}$, find the nearest $\mathbf{T_{ba}}\mathbf{b_j}$ from all $j$, denoted as $\mathbf{b_{i^*}}$. %\triangleq f_x(i)$.
\item For each $\mathbf{b_j}$, find the nearest $\mathbf{T_{ab}}\mathbf{a_i}$ from all $i$, denoted as $\mathbf{a_{j^*}}$. %\triangleq f_y(j)$.
\item Optimize $\mathbf{T_{ab}}$ and $\mathbf{T_{ba}}$ by minimizing:
\begin{equation}
\begin{aligned}
%&\sum_{j} || y_j - T_{xy}x_{f_y(j)} ||_2 + \sum_{i} || x_i - T_{yx}y_{f_x(i)} ||_2 \\
L_{trans} = & \sum_{i} ||\mathbf{b_{i^*}} - \mathbf{T_{ab}}\mathbf{a_i} ||_2 +  \sum_{j} ||\mathbf{a_{j^*}} - \mathbf{T_{ba}}\mathbf{b_j} ||_2  \\ %這和上一行是不是等價的？如果 x 和 y 最近，y 和 x 也會最近 -- Lee
&+ \lambda^\prime \sum_{i} || \mathbf{a_i} - \mathbf{T_{ba}}\mathbf{T_{ab}}\mathbf{a_i} ||_2 \\
&+ \lambda^\prime \sum_{j} || \mathbf{b_j} - \mathbf{T_{ab}}\mathbf{T_{ba}}\mathbf{b_j} ||_2
\end{aligned} \label{eq:align}
\end{equation} 
\end{enumerate}
%\lambda ->　\lambda^\prime　(因為 lambda 前面有了)
In the first and the second terms, we want to transform $\mathbf{a_i}$ and $\mathbf{b_j}$ respectively to its nearest neighbors  in the other space, $\mathbf{b_{i^*}}$ and $\mathbf{a_{j^*}}$.
We include cycle-constraints as the third and fourth terms in (\ref{eq:align}) to ensure that both $\mathbf{a_i}$ and $\mathbf{b_j}$ are unchanged after transformed to the other space and back. 

Equation (\ref{eq:align}) is solved by gradient descent.
After $\mathbf{T_{ba}}$ is eventually obtained, given $\mathbf{a_i}$, we can find $\mathbf{b_{i^*}}$ in which $\mathbf{T_{ba}}\mathbf{b_{i^*}}$ is nearest to $\mathbf{a_i}$ among all the columns of $B$.
Then we consider the $i^*$-th word in the text database corresponds to the $i$-th audio segment, or the $i^*$-th word is the recognition result of the $i$-th audio segment.

If some aligned pairs of audio and textual semantic embeddings are available, we can also train the transformation matrix in a supervised/semi-supervised way, in which we directly minimize the distance from the true  embedding in the first two terms of (\ref{eq:align}), rather than from the nearest embedding. %我這樣寫對嗎？　-- Lee

%Procrustes method 不是 supervised 的嗎？先註解掉，如果真的有用到，應該寫清楚 -- Lee
%We also use Procrustes method~\cite{xing2015normalized} to fine-tune the transformation matrix. 
%It solves a supervised equation of the form $\min_{\mathbf{W}} || \mathbf{WX} - \mathbf{Y} ||_2$ where $\mathbf{W}$ is orthonormal. 
%Suppose the decomposition of $\mathbf{YX^T}$ is $\mathbf{U}\sum\mathbf{V^T}$. 
%The optimal $\mathbf{W}$ given by the Procrustes method is recovered by $\mathbf{UV^T}$.

\begin{table}[t]
\footnotesize
\centering
\caption{The Spearman's rank correlation scores of four audio sets with textual semantic embeddings with one-hot input, where the abbreviations are \textbf{SE}: semantic embeddings, \textbf{PE}: phonetic embeddings, \textbf{SAD}: sequence-to-sequence autoencoder with disentanglement, \textbf{SA}: sequence-to-sequence autoencoder without disentanglement}
\label{table:one-hot}
\begin{tabular}{|c|c|c|c|c|}
\hline
%\multicolumn{1}{|c|}{\multirow{2}{*}{\textbf{Dataset}}} & \multicolumn{2}{c|}{\textbf{disentangle}} & \multicolumn{2}{c|}{\textbf{no disentangle}}  \\ \cline{2-5} 
\multicolumn{1}{|c|}{\textbf{Dataset}} & \multicolumn{1}{c|}{\textbf{SE/SAD}} & \multicolumn{1}{c|}{\textbf{SE/SA}} & \multicolumn{1}{c|}{\textbf{PE/SAD}} & \multicolumn{1}{c|}{\textbf{PE/SA}}     \\ \hline \hline
%AP         & 0 & 0 & 0 & 0 \\  
%BLESS      & 0 & 0 & 0 & 0 \\
%Battig     & 0 & 0 & 0 & 0 \\
%ESSLI\_1a  & 0 & 0 & 0 & 0 \\
%ESSLI\_2b  & 0 & 0 & 0 & 0 \\
%ESSLI\_2c  & 0 & 0 & 0 & 0 \\ \hline
MEN~\cite{bruni2014multimodal}        & \bf{0.415} & 0.149 & 0.297 & 0.073 \\  
MTurk~\cite{radinsky2011word}      & \bf{0.442} & 0.251 & 0.373 & 0.221 \\
RG65~\cite{rubenstein1965contextual}       & 0.236 & 0.217 & \bf{0.273} & -0.073 \\
RW~\cite{luong2013better}         & \bf{0.730} & 0.595 & 0.684 & 0.550 \\
SimLex999~\cite{hill2015simlex}  & \bf{0.309} & -0.043 & 0.139 & -0.110 \\
WS353~\cite{finkelstein2001placing,agirre2009study}      & \bf{0.441} & 0.203 & 0.374 & 0.109 \\
WS353R~\cite{finkelstein2001placing,agirre2009study}     & \bf{0.385} & 0.164 & 0.348 & 0.102 \\
WS353S~\cite{finkelstein2001placing,agirre2009study}     & \bf{0.465} & 0.224 & 0.367 & 0.122 \\ \hline
% Google     & 0 & 0 & 0 & 0 \\
% MSR        & 0 & 0 & 0 & 0 \\
% SemEval    & 0 & 0 & 0 & 0 \\ \hline
\end{tabular}
\end{table}

\begin{table}[t]
\footnotesize
\centering
\caption{The Spearman's rank correlation scores of four audio sets with textual semantic embeddings with phonetic embedding input.}
\label{table:SA}
\begin{tabular}{|c|c|c|c|c|}
\hline
%\multicolumn{1}{|c|}{\multirow{2}{*}{\textbf{Dataset}}} & \multicolumn{2}{c|}{\textbf{disentangle}} & \multicolumn{2}{c|}{\textbf{no disentangle}}  \\ \cline{2-5} 
\multicolumn{1}{|c|}{\textbf{Dataset}} & \multicolumn{1}{c|}{\textbf{SE/SAD}} & \multicolumn{1}{c|}{\textbf{SE/SA}} & \multicolumn{1}{c|}{\textbf{PE/SAD}} & \multicolumn{1}{c|}{\textbf{PE/SA}}     \\ \hline \hline
%AP         & 0 & 0 & 0 & 0 \\  
%BLESS      & 0 & 0 & 0 & 0 \\
%Battig     & 0 & 0 & 0 & 0 \\
%ESSLI\_1a  & 0 & 0 & 0 & 0 \\
%ESSLI\_2b  & 0 & 0 & 0 & 0 \\
%ESSLI\_2c  & 0 & 0 & 0 & 0 \\ \hline
MEN        & \bf{0.430} & 0.303 & 0.381 & 0.207 \\  
MTurk      & 0.471 & \bf{0.492} & 0.371 & 0.373 \\
RG65       & \bf{0.161} & 0.152 & 0.117 & 0.003 \\
RW         & \bf{0.712} & 0.670 & 0.696 & 0.585 \\
SimLex999  & \bf{0.273} & 0.249 & 0.221 & 0.129 \\
WS353      & \bf{0.520} & 0.478 & 0.504 & 0.393 \\
WS353R     & \bf{0.525} & 0.463 & 0.470 & 0.406 \\
WS353S     & 0.502 & 0.501 & \bf{0.526} & 0.381 \\ \hline
% Google     & 0 & 0 & 0 & 0 \\
% MSR        & 0 & 0 & 0 & 0 \\
% SemEval    & 0 & 0 & 0 & 0 \\ \hline
\end{tabular}
\end{table}

%%%%%%%%% EXPERIMENTS
\section{Experiments}

%0. setup and dataset \\
\subsection{Experimental Setup}
\label{subsec:setup}

We used LibriSpeech~\cite{panayotov2015librispeech} as the audio collection in our experiments.
% * <sungsung3201@gmail.com> 2018-03-23T18:04:18.968Z:
% 
% 所有LibriSpeech的S都要改成大寫
% 
% ^.
LibriSpeech is a corpus of read English speech and suitable for training and evaluating speech recognition systems. It is derived from audiobooks that are part of the LibriVox project, and contains 1000 hours of speech sampled at 16 kHz. 
In our experiments, the dataset were segmented according to the word boundaries obtained by forced alignment with respect to the reference transcriptions.
We used the 960 hours "clean+others" speech sets from LibriSpeech for model training.
MFCCs of 39-dim were used as the acoustic features. 

The phonetic encoder, speaker encoder and decoder in the first stage are all 2-layer GRUs with hidden layer size 256, 256 and 512, respectively.
The discriminator is a fully-connected feedforward network with 2 hidden layer whose size is 256.
The $\lambda$ value we used in the speaker loss term is set to 0.01.
The discriminator and the other parts of this stage are iteratively trained as WGAN~\cite{gulrajani2017improved}.

In the second stage, the two encoders are both 2-hidden-layer fully-connected feedforward networks with hidden size 256. 
The size of embedding vectors is 128, the context window size 5, the negative sampling number 5, the sampling factor 0.001, and 5 as the threshold for minimum count.　
Although textual Word2Vec has an unsupervised training procedure, it needs subsampling, which is an important step during training.
Subsampling needs the frequencies of words, which we can't obtain during the unsupervised training in the audio domain.
In this preliminary work, we compromise to use known word labels to do subsampling, but we believe, with a proper statistically clustering algorithm to classify and group phonetic vectors, a completely unsupervised ASR can be achieved by this framework in the future.

We trained two sets of textual semantic embeddings.
The first set (denoted as OHW in the following discussion) was trained on the manual transcriptions of LibriSpeech using one-hot representations as input by a typical skip-gram model.
% * <sungsung3201@gmail.com> 2018-03-23T17:59:14.299Z:
% 
% 應該講出簡寫怎麼來的
% 例如：
% The first set, One-Hot-as-input Word vectors (denotes as OHW in the following discussion), was trained...
% 
% ^.
The second set of textual semantic embeddings (denoted as PEW) was also trained on LibriSpeech while using phonetic information.
% * <sungsung3201@gmail.com> 2018-03-23T18:01:24.473Z:
% 
% 也一樣要講PEW是什麼的簡寫，例如
% The second set of textual semantic embeddings, Phoneme-encoded-as-input Word vectors (denoted as PEW), was also...
% 
% ^.
To generate PEW, we represented each word with a sequence of phonemes, and used a sequence-to-sequence autoencoder to encode the phoneme sequence into an embedding with size of 256.
Then we took the embeddings of phoneme sequences as input of skip-gram model. %as in audio semantic training.
%In this way, the textual semantic embeddings also encoded phonetic information.
%as audio semantic embeddings, which can help the transformation from audio semantic embeddings to textual semantic embeddings in the next stage.
Because the audio semantic embeddings were also learned from phonetic embeddings, we believe PEW would have a more similar distribution to audio semantic embeddings than OHW.

%====================== 我這樣寫符合事實嗎？ -- Lee
While each word in text has a unique semantic representation, segments corresponding to the same word can have different semantic representations $\mathbf{v_w}$, so the distributions of textual and audio semantic embedding are too different to be mapped together. 
In this work, we used known word labels to average audio semantic embeddings $\mathbf{v_w}$ corresponding to the same word, so that each word has a unique semantic representation in both audio and text.
We realize that this is another unrealistic setup in unsupervised scenario, and will develop technique to address this issue in the future.
%====================== 未來真的要解決這個問題
Finally in the third stage, we applied MBC-ICP~\cite{hoshen2018iterative} with top 5000 frequent words and projected the embeddings to the top 100 principle components. Hence, the affine transform matrix from audio embeddings to text embeddings is $100 \times 100 $, and vise versa. The mini-batch size was set to be 200.
%However, because the audio segments of the same word have different signals, and thus their phonetic embeddings are not exactly the same, the semantic embedding of the same word can be different. 
%Hence, the mapping from audio space to textual space is a  many-to-one mapping, and a well-designed algorithm is needed for unsupervised ASR. %challenging
%Again, to alleviate the clustering problem, we use the word labels of audio semantic embeddings to average the embedding vectors corresponding to the same word. % 如果有這麼做，也留待實驗再講

The models in the first two stages were trained with Adam optimizer~\cite{kingma2014adam}, while the last stage was trained using stochastic gradient descent with the initial learning rate 0.01, which decayed every 40 iterations with rate 0.90.

%2-1. semantic vectors trained from phonetic/undisentangled vectors & simply phonetic/undisentangled vectors. \\
\subsection{Evaluation of Word Representations}

%Many papers have researched on word representations and propose many testing word pairs for word vector representations~\cite{jastrzebski2017evaluate,radinsky2011word,bruni2014multimodal,finkelstein2001placing,agirre2009study,rubenstein1965contextual,luong2013better,hill2015simlex}.
%這幾句是多餘的
%Many of them employ a transfer learning view, in which the main goal of representation learning is to make subsequent learning fast, i.e. use resulting word embeddings to maximize performance at the lowest sample complexity possible.
%It is crucial to report model (using given word representation) performance under varying (benchmark) dataset sizes and model classes, which is important for correct evaluation of transfer.
%Based on it, 
Several benchmarks for evaluating  word representations are available~\cite{jastrzebski2017evaluate,radinsky2011word,bruni2014multimodal,finkelstein2001placing,agirre2009study,rubenstein1965contextual,luong2013better,hill2015simlex}.
Those benchmark corpora include pairs of words.
Here we want to know whether the audio semantic embeddings can capture semantic meanings like textual semantic embeddings.
We calculated cosine similarities of word pairs  in benchmark datasets by audio semantic embeddings and text semantic embeddings, and evaluated the ranking correlation scores of cosine similarities between audio and textual embeddings\footnote{The code is released at https://github.com/grtzsohalf/Towards-Unsupervised-ASR}.
The results are presented with the Spearman's rank correlation coefficients.

%The first block in the table is used to measure the categorization performance.
%The values in the table represent cluster purity of embedding vectors.
%The second block is used to evaluate the similarity between human-defined scores of word pairs and cosine similarities of embedding vectors corresponding to the word pairs.
%The bottom sets in the table are used to test analogy of words.
%The values are analogy prediction accuracies.

We measured correlations between two textual semantic embeddings, OHW and PEW, mentioned in Section~\ref{subsec:setup} and four types of audio embeddings.
The four  types of audio embeddings are: semantic embeddings trained from phonetic vectors with disentanglement (SE/SAD), semantic embeddings trained from vectors extracted by  SA without disentanglement (SE/SA), phonetic embeddings extracted by SA with disentanglement (PE/SAD), and embeddings extracted by   SA without disentanglement (PE/SA).

The results are shown in Table~\ref{table:one-hot} and Table~\ref{table:SA}.
Table~\ref{table:one-hot} shows the correlations between OHW and four audio embedding sets. 
Similarly, Table~\ref{table:SA} presents the correlations between PEW and four audio embedding sets.
%We can observe that the audio semantic embedding set with disentanglement achieved the best performance among four audio embeddings in both tables, and 
In Tables~\ref{table:one-hot} and~\ref{table:SA}, we found that disentanglement improved the correlation scores in most  cases, and audio semantic embeddings outperformed embeddings extracted from SA in most cases.
The results verify the first two stages in our proposed method are both helpful for extracting embeddings including semantic information.
It can also be inferred from the two tables that correlation performance of PEW is better than OHW as expected because PEW is learned from the phonetic embeddings of text.
Since PEW is more similar to the audio embeddings, it will make the transformation in the next stage easier. 
%because phonetic information in text can actually make audio semantic embeddings more similar to textual semantic embeddings.

\begin{table}[t]
\footnotesize
\centering
\caption{Comparison of top 10 nearest transformation accuracy between SE/SAD, SE/SA, PE/SAD and PE/SA.}
\label{table:ten}
\begin{tabular}{|c|c|c|c|c|}
\hline
\multicolumn{1}{|c|}{\textbf{Labeled Pairs }} & \multicolumn{1}{c|}{\textbf{SE/SAD}} & \multicolumn{1}{c|}{\textbf{SE/SA}} & \multicolumn{1}{c|}{\textbf{PE/SAD}} & \multicolumn{1}{c|}{\textbf{PE/SA}}   \\ \hline \hline
5000          & 0.6068   & 0.3902  & 0.4736 & 0.4744\\
\hline 
\end{tabular}
\end{table}

\begin{table}[t]
\footnotesize
\centering
\caption{Top K nearest transformation accuracies of SE/SAD using semi-supervised methods with different numbers of labeled pairs.}
\label{table:k}
\begin{tabular}{|c|c|c|c|}
\hline
\multicolumn{1}{|c|}{\textbf{Labeled Pairs }} & \multicolumn{1}{c|}{\textbf{top 1}} & \multicolumn{1}{c|}{\textbf{top 10}} & \multicolumn{1}{c|}{\textbf{top 100}}    \\ \hline \hline
0             &  0.000  & 0.0014   &  0.0194 \\ 
1000          &  0.0400 &  0.1378  & 0.4984  \\
2000          &  0.1080  & 0.4526 &  0.7578 \\
5000          &  0.1846  & 0.6068 &  0.8638 \\
\hline 
\end{tabular}
\end{table}

% \begin{table}[]
% \footnotesize
% \centering
% \caption{The top 10 nearest audio-to-text transformation results of some sampled words.}
% \label{table:words}
% \begin{tabular}{|c|c|c|}
% \hline

% this         & that,this,there,f,i've,has,either,as,her,and     \\
% \hline
% would & what,went,one,won't,will,would,want,don't,wood,we'll      \\
% \hline
% best & best,blessed,rest,lest,lost,laughed,once,life,writes,liked \\
% \hline
% my   & my,i'm,i've,i'd,i'll,run,van,gone,not,aunt \\
% \hline 
% \end{tabular}
% \end{table}

\subsection{Transformation from Audio to Text}

%It is worth noting that, phonetic embedding input outperforms one-hot input because it contains phonetic information.
%To examine whether there is a good affine transformation matrix, we adopted supervised training at first, and then apply MBC-ICP in an unsupervised way.
% Then we gradually reduce the number of labelled audio and textual semantic embedding pairs to see if the transformation matrix from audio embeddings to textual semantic embeddings could still be obtained. 
The results of MBC-ICP are shown in Table~\ref{table:ten} and Table~\ref{table:k}. 
In Table~\ref{table:ten}, we compare top 10 nearest accuracies of four audio semantic sets mentioned above using 5000 labeled pairs. %using supervised learning.
SE/SAD achieved the best result.
Once again, it shows that the first two stages in our proposed method are both effective indeed.
In Table~\ref{table:k}, both the unsupervised and semi-supervised results with SE/SAD are further reported. %of SE/SAD
%The audio embeddings used in Table~\ref{table:k} are SE/SAD. 
%For semi-supervised learning, we adopted supervised training on labeled data at first, and then applied MBC-ICP in an unsupervised way. %<-前面 2.3 應該會解釋吧，這邊就不用講了對吧？-- Lee
The numbers of labeled data are 0, 1000, 2000 and 5000 respectively.
The results also include top 1, top 10 and top 100 nearest accuracies.
We can observe that although unsupervised MBC-ICP may not generate perfect matching, semi-supervised learning achieved high transformation accuracies. 
It shows that there  exists a good affine transformation matrix that can transform audio and textual semantic embeddings.
However, the affine transformation matrix cannot be easily found by the completely unsupervised approach.
%======================== 如果空間不夠的話，這一段感覺可以優先刪掉，因為結果沒有非常驚人 -- Lee
% To further analyze the transformation results, we randomly sampled some words from SE/SAD, and list the top ten nearest tranformation results in Table~\ref{table:words}.
% The results show that audio semantic embeddings actually embed both semantic and phonetic information. %not only contain phonetic information but also embed their semantic meanings.
% %And after the transformation from audio to text, these information can still be well preserved.
% Taking the word, ``would'', as an exmple, in the top ten results, ``what'', ``went'', ``want'', ``wood'' are phonetically similar while ``won't'', ``will'', ``don't'', ``we'll'' are semantically related.
%========================

%%%%%%%%% CONCLUSIONS AND FUTURE WORK
\section{Conclusions and Future Work}

In this work, we propose a three-stage framework towards unsupervised ASR with unaligned speech and text only.
Through the experiments, we showed semantic audio embeddings can be directly extracted from audio. %prove that suitable disentangled phonetic embeddings extracted from audio segments can effectively preserve their phonetic structures. 
Although we did not obtain satisfied results with unsupervised learning, via semi-supervised learning, we verified there  is an affine matrix transforming semantic embeddings from audio to text.
How to find the affine matrix in an unsupervised setup is still under investigation.
%To achieve unsupervised transformation from audio to text, we propose an approach to produce audio semantic embeddings and prove there is an affine matrix transforming semantic embeddings from audio to text.
Although some oracle settings were used in the experiments, we are conducting experiments under more realistic setups.
We believe with further improvement on this framework, the completely unsupervised ASR could be achieved in the near future. 

% \section{Acknowledgements}

% Something may be useful: \\

% 5\textsuperscript{th}

% \begin{itemize}
% \item Proceedings will be printed in DIN A4 format. Authors must submit their papers in DIN A4 format.
% \end{itemize}

% \begin{enumerate}
% \item For Windows users, the bullzip printer can convert any PDF to have embedded and subsetted fonts.
% \item For Linux/Mac users, you may use \\
%    pdftops file.pdf\\
%    pstopdf -dPDFSETTINGS=/prepress file.pdf
% \end{enumerate}

% e.\,g.\ ``http://www.foo.com/index.html'' \\
% \emph{Don't make any footers or headers!} \\

%%%%%%%%% REFERENCES

\bibliographystyle{IEEEtran}

\bibliography{mybib,IR_bib,ref_dis,segment,transfer,INTERSPEECH16,ICASSP13}

\end{document}